\documentclass{new_tlp}
\usepackage{mathptmx}

\newcommand\bcmdtab{\noindent\bgroup\tabcolsep=0pt%
  \begin{tabular}{@{}p{10pc}@{}p{20pc}@{}}}
\newcommand\ecmdtab{\end{tabular}\egroup}

\usepackage{graphicx}
\usepackage{url}

\usepackage{subcaption}
\usepackage{MnSymbol}
\usepackage{multicol}

\usepackage{algorithm}
\usepackage{algorithmic}
\usepackage{wrapfig}
\usepackage{rotating}
\usepackage{tikz}
\usepackage{adjustbox}
\usepackage{blindtext}
\usepackage{hvfloat,lipsum}
 
\newcommand{\beitemize}{\begin{list}{$\bullet$}{\topsep=1.5pt \parsep=0pt \itemsep=1pt \leftmargin=1em }} 
\newcommand{\enitemize}{\end{list}}

\newcommand{\beenumerate}{\hspace{-0.5in} \begin{enumerate}\topsep=1pt \parsep=0pt \itemsep=-3pt} 
\newcommand{\enenumerate}{\end{enumerate}}

\usepackage{picinpar}
\def\naf{\; not \; } 
\def\asposs{{\sc asp4OSS}}

\makeatletter

\renewcommand{\section}{\@startsection
{section}%
{1}%
{\z@}%
{-.5\baselineskip}%
{0.4\baselineskip}
{\centering \bf\boldmath\pretolerance=10000\relax}}
 
\renewcommand{\subsection}{\@startsection
{subsection}%
{2}%
{\z@}%
{-.5\baselineskip}%
{0.4\baselineskip}
{\centering \boldmath\bf\it\pretolerance=10000\relax }}%

\renewcommand{\subsubsection}{\@startsection
{subsubsection}%
{2}%
{\z@}%
{0.3\baselineskip}%
{0.4\baselineskip}
{\centering \em \pretolerance=10000\relax }}%

\makeatother

  \title[Explaining the Three Mile Island Nuclear Accident Scenario]
        {An Application of ASP in Nuclear Engineering: 
        Explaining the Three Mile Island Nuclear Accident Scenario}

  \author[B. N. Hanna, L. T. Trieu, T. C. Son, and N. T. Dinh]
         {BOTROS N. HANNA, LY LY T. TRIEU, TRAN C. SON\\ 
            Computer Science Department, New Mexico State University, Las Cruces, New Mexico, USA\\
         \email{\{bn,lytrieu,stran\}@nmsu.edu}
          \and NAM T. DINH\\
            Department of Nuclear Engineering, North Carolina State University, Raleigh, North Carolina, USA\\
         \email{ntdinh@ncsu.edu}
         }         

\jdate{March 2003}
\pubyear{2003}
\pagerange{\pageref{firstpage}--\pageref{lastpage}}
\doi{S1471068401001193}

\begin{document}

\label{firstpage}

\maketitle

  \begin{abstract}
The paper describes an ongoing effort in developing a declarative system for supporting operators in the Nuclear Power Plant (NPP) control room. The focus is on two modules: diagnosis and explanation of events that happened in NPPs. We describe an Answer Set Programming (ASP) representation of an NPP, which consists of declarations of state variables, components, their connections, and rules encoding the plant behavior. We then show how the ASP program can be used to explain the series of events that occurred in the Three Mile Island, Unit 2 (TMI-2) NPP accident, the most severe accident in the USA nuclear power plant operating history. We also describe an explanation module aimed at addressing answers to questions such as ``why an event occurs?'' or ``what should be done?'' given the collected data. 

This paper is *under consideration* for acceptance in TPLP Journal.
  \end{abstract}



\section{Introduction} 
\label{sec:introduction}  

When an incident occurs in the Nuclear Power Plant (NPP), the operator is expected to diagnose faults and make decisions promptly to keep the plant operational and safe. A misdiagnosis or a late decision by the operator may cause catastrophic accidents. Although operators go through years of training before working in an NPP control room, 80\% of the NPP incidents are attributed to human error~\cite{standard2009human}. 

The occurrence of human error is due to the complexity of the system that produces many alarms after a malfunction in addition to the psychological pressure that the operator goes through. Besides, it is overwhelming to monitor hundreds of indicators in the control room and interact with dynamic events. If proper control actions are not executed on time, nuclear reactor trip set points are reached, and a reactor shutdown cannot be avoided. NPPs can lose over \$1,000,000 in revenue for every day they are shut down~\cite{lew2018transitioning}.

Since human error may impact NPP safety and economic viability, it is a prominent goal of the nuclear industry to minimize the chances of human error. Therefore, there has been an increased interest in the implementation of AI methods that integrate information across the control room to assist the operator in diagnosing the faults and making corrective actions. Additionally, these AI methods may minimize the needed staffing size and enable deploying reactors at remote sites.


Recently, there have been numerous efforts toward implementing AI methods that are capable of diagnosis, identifying viable actions, and recommending decisions to the operator. AI statistical methods (such as deep learning and Bayesian network) that support the operator decisions have been proposed~\cite{darling2018intelligent,boroushaki2003intelligent,peng2018intelligent}. Statistical methods assume that big data from the NPP history or simulations are available to predict unknown variables (e.g., pipe break size or a component failure probability). The success of these methods relies on the sufficiency and relevance of the available data. At the same time, the reliance on simulated data, the sensitivity with changes in data, and the ''black-box'' style of these methods provide a real challenge for their deployment. 

Because of the issues of statistical approaches, an alternative has been developed to support operators of the NPP. In this approach, the NPP system is described as a knowledge base, i.e., a set of facts, logic rules, and constraints. Examples of these rules may include flow paths, operation constraints, and emergency procedures. Automated reasoning is then used with the knowledge base to identify plans or diagnoses which support the operators' decision-making process.  Some of the proposed reasoning methods to support the NPP operators are implemented using Java-based rules engines such as the online fault detection system~\cite{reifman1999prodiag,park2017implementation} and the multilevel model for reasoning about causes and consequences of malfunctions~\cite{lind2014functional}. Other diagnosis methods \cite{chang1995development} are based on the logic programming language, Prolog \cite{clocksin2012programming}. 

In \cite{hanna2019artificial}, we proposed an Answer Set Programming (ASP) based system for supporting the NPP management and demonstrated the capability of the system by computing a plan for the operator to deal with situations such as a turbine control valve that drifts stuck closed when it is supposed to be open. \color{black}

Compared to other knowledge representation and reasoning methods \cite{reifman1999prodiag,park2017implementation,lind2014functional,chang1995development} that have been proposed to support the operators' decision-making process, ASP has some attractive features. ASP is fully declarative and more flexible compared to Prolog, e.g., it can deal with a program containing negative cycles while Prolog will need to use a special technique for it. Additionally, the ASP-based reasoning system proved to have better performance and availability compared to the Java-based rules engines \cite{liang2009openrulebench}. Besides, none of the previous reasoning-based NPP operator support systems were applied to or tested against a realistic complex scenario such as the TMI-2 accident. \color{black}


This paper is a continuation of earlier work \cite{hanna2019artificial}.  The objectives of this paper are: 
\beenumerate
\item Testing the capability of an ASP-based reasoning system by applying it to a real-world situation with significant consequences.    
\item Demonstrating that the ASP-based reasoning system answers (diagnoses and recommendations) can be accompanied by rational explanations.
\enenumerate
To achieve the above goals, we experiment with the ASP-based NPP operator support system with the available data about the most severe nuclear accident in US history, the Three Mile Island, Unit 2 (TMI-2) accident \cite{nuclear1979investigation,nuclear1980analysis}. We experiment with this accident for different reasons. First, a large amount of data about system behavior prior to the accident is available. Second, this data has been analyzed extensively by nuclear engineering researchers, which identify precisely the reasons of the accident as well as possible recommendations that could help to avoid the accident. Third, the data has not been looked at from the perspective of a declarative reasoner, such as the NPP operator support system. To the best of our knowledge, this is the first work to apply an automated reasoning system to the TMI-2 accident scenario.

\section{Background} 

\subsection{The Nuclear Power Plant System}
\label{sec:npp}

\begin{figure} 
    \centering
    \includegraphics[width=1\textwidth]{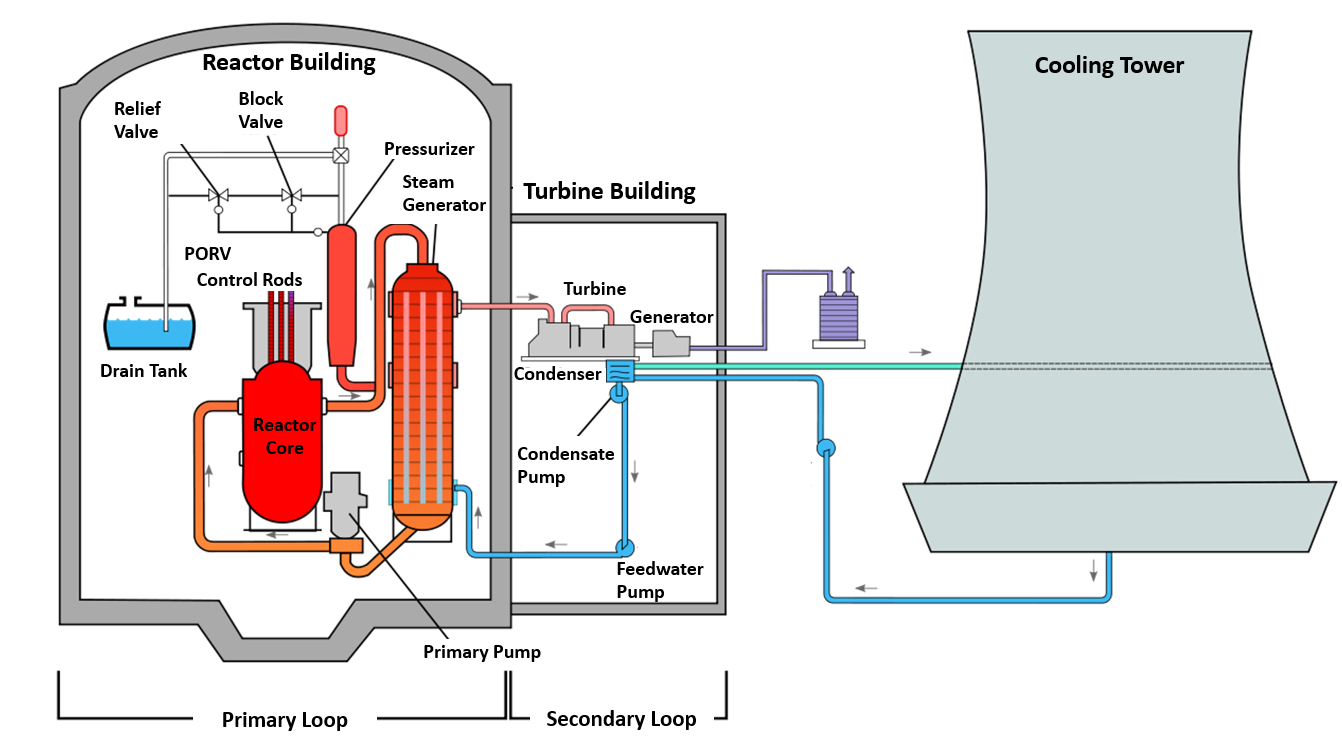}
    \caption{A diagram of the TMI-2 NPP \protect\cite{WinNT}}
  \label{fig:overall_system}
\end{figure}

A simplified view of the TMI-2 reactor system is depicted in Figure~\ref{fig:overall_system}. \color{black} The reactor design depicted in Figure~\ref{fig:overall_system} is called `pressurized water reactor' which  constitute most of the world plants today. \color{black} 

In the \emph{reactor core}, a self-sustained nuclear fission reaction is controlled using the \emph{control rods} that absorb neutrons available for fission. If the control rods are inserted deeper into the reactor, the heat created by the reactor and the reactor power output decrease and vice versa. The reactor coolant is pressurized (kept under high pressure) water, so it does not boil. Pressurized water is pumped, through the \emph{primary loop}, by the \emph{primary pump}, through the reactor and absorbs heat from the nuclear fuel. 

The reactor coolant goes through the  \emph{steam generator}, where heat is transferred to lower-pressure water in the \emph{secondary loop} (note that primary and secondary loops' coolants do not mix). Secondary loop low-pressure water is allowed to boil to produce steam for the \emph{turbine}, causing it to turn the \emph{generator} that produces electricity. The unused steam is exhausted to the  \emph{condenser} to get condensed into water that is pumped back, by the \emph{condensate pump} and the \emph{feedwater pump}, to the steam generator. Auxiliary (emergency) feedwater pumps are available to supply water to the steam generator to remove heat from the reactor when the reactor is shutdown.

One of the major components in the primary loop is the \emph{pressurizer}, whose function is maintaining the primary coolant pressure by a heater and water sprayer to increase/decrease the coolant pressure. If needed, valves are used to release any excess pressure. Some of these valves are shown in Figure~\ref{fig:overall_system}: the \emph{Pilot Operated Relief Valve (PORV)} and the \emph{block valve}.

For simplicity, Figure~\ref{fig:overall_system} shows only one primary loop and one secondary loop. However, the TMI-2 Reactor Coolant System (RCS) consists of 2 flow loops (loops A and B), each with a steam generator, a reactor, and pumps. Both flow loops (A and B) share the same pressurizer.

Before the accident, a leakage, accumulated in the reactor coolant \emph{drain tank}, was identified to be from the PORV and was calculated to be within technical limits. Additionally, the secondary loop auxiliary feedwater line block valves, that are normally open, were inadvertently left closed ~\cite{nuclear1979investigation}. An overview of the TMI-2 accident is presented in the next section.

\subsection{A Summary of the Three Mile Island Accident}
\label{sec:history} 

The TMI-2 partial reactor meltdown on March 28th, 1979, is the most severe accident in the NPP history in the US. The TMI-2 accident occurred because of equipment failures, human errors, design-related problems, and lack of training. However, we present a simplified summary of the first 142 minutes of the accident based on a view that focuses on human errors and human reasoning. The accident can be summarized as follows \cite{nuclear1979investigation,nuclear1980analysis}: 

The accident started when the plant experienced a loss of main feedwater and the resulting automatic shutdown of the turbine. The auxiliary feedwater pumps started automatically, as designed, to make up for the loss of feedwater flow. Within the first minute of the event, the steam generator water level dropped to the point where automatic controls called for the auxiliary feedwater, to maintain a minimum steam generator water level. However, closed block valves between the control valves and the steam generators initially prevented the auxiliary feedwater from being delivered to the steam generators. These valves were opened by the operators 8 minutes after the accident was initiated.

Immediately after the loss of main feedwater, the RCS pressure began to increase. When the RCS pressure increased to 2255 PSI, the pressurizer PORV opened to relieve the excess pressure, but RCS pressure continued to increase, and the reactor was tripped automatically. With the reactor trip, the RCS pressure decreased, but the pressurizer PORV failed to close when the closure setpoint was reached, and the coolant continued to escape through the PORV whose failure to close was not recognized by the staff for some time. As the escape of water continued, the water level in the
system fell. The pressure in the pressurizer also fell, falling so low that the generation of steam started in the reactor core, a condition for which this type of reactor is not designed (operators did not know about the formation of steam in the RCS). 

Because the RCS pressure continued to decrease, an emergency cooling system, the High-Pressure Injection System (HPIS), started automatically at about 2 minutes. Therefore, the pressurizer water level was rising until the water level was off the scale, which indicated that the pressurizer (and the whole RCS) is expected to be filled with water or ``going solid.'' The operators were trained not to let the pressurizer get solid. Therefore, the operators decided to throttle the HPIS pumps, but the pressurizer water level kept rising.

As reactor coolant inventory decreased because of the stuck open pressurizer PORV, the primary pumps’ output flow rate decreased until the primary pumps were finally stopped. After more than 2 hours of the beginning of the accident, the PORV block valve was closed. Because the primary loop temperature continued to rise, the operators decided to turn on the HPIS pumps to get the reactor refilled with water. Finally, the RCS was filled with water, but the reactor core was partially melted. The sequence of events is presented in Table \ref{events}.

\begin{table}
\caption{The TMI-2 accident key sequence of events}\label{events}
\begin{tabular}{lll}
\hline\hline
Clock time (AM)     & Time(seconds)     & Event\\
\hline
04:00:36    & 0     & Normal operation\\
04:00:37    & 1     & Condensate pumps trip\\
04:00:38    & 2     & Feedwater pumps trip\\
04:00:38    & 2     &Turbine trips\\
04:00:38    & 2     & Auxiliary  feedwater pumps start\\
04:00:43    & 7     & Pressuirzer PORV opens due to high primary loop pressure\\
04:00:47    & 11     & Reactor trips\\
04:00:47    & 11     & Primary loop pressure starts decreasing, but the PORV remains open.\\
04:02:38    & 122     & HPIS system starts\\
04:05:15    & 279     & Operators throttle HPIS pumps\\
04:08:55    & 499     & The auxiliary  feedwater line block valve is opened (loop A)\\
04:08:56    & 500     & The auxiliary feedwater line block valve is opened (loop B)\\
05:13:58    & 4402     & Primary pump is tripped (loop A)\\
05:41:12    &6036     & Primary pump is tripped (loop B)\\
06:22:37    &8521     & Pressurizer block valve is closed\\
\hline\hline
\end{tabular}
\end{table}

In this complex scenario, it was challenging for the operators to make timely decisions due to:
\beitemize 
\item \emph{A large number of time-dependent variables}: For instance, many alarms occurred, and the operator did not notice the containment sump high water level alarm (which suggested leakage from the RCS). Also, the operators were concerned about an apparently rising primary water level, and did not notice that the primary loop water reached saturation pressure (water converted to steam) \cite{results}, The pressurizer water level was high because of a growing steam region in the core.
\item \emph{The misleading reading}: One of the complications of the TMI-2 scenario is the fact that the pressurizer PORV was stuck open. The operators believed that it was closed because of the control room indication, but the PORV status light was just an indication that the 
PORV solenoid de-energized, giving a non-open indication to the control room operator.  The PORV should have reseated, but did not.
\enitemize 

\subsection{Answer Set Programming} 
\label{subsection:asp} 

\emph{Answer set programming} 
(ASP)~\cite{MarekT99,Niemela99} is a declarative programming paradigm based on 
logic programming under the answer set semantics.   
A logic program $\Pi$ is a set of rules of the form  \quad 
 {$c  \leftarrow a_1,\ldots,a_m,\naf b_{1},\ldots,\naf b_n$} \quad
where $c$, $a_i$'s, and $b_i$'s are atoms of a propositional language\footnote{For simplicity, we often use first order logic atoms in the text which represent all of its ground instantiations.} and 
$\mathit{not}$ represents (default) negation. 
Intuitively, a rule  states that if $a_i$'s are  believed to be true and none of the $b_i$'s is believed to be true then $c$ must be true. For a rule $r$, 
$r^+$ and $r^-$   denote the sets $\{a_1,\ldots,a_m\}$ and $\{b_{1},\ldots,b_n\}$, respectively. \color{black} We write $head(r)$ to denote $c$ and refers to the right side of the rule as the body of $r$. \color{black}
 
Let $\Pi$ be a program. An interpretation $I$ of $\Pi$ is a set of ground atoms occurring in $\Pi$.
The body of a rule $r$ is satisfied by $I$ if $r^+ \subseteq I$ and $r^- \cap I = \emptyset$.
A rule $r$ is satisfied by $I$ if the body of $r$ is satisfied by $I$ implies $I \models c$.  
When $c$ is absent, $r$ is a constraint and is satisfied by $I$ if its body is not satisfied by $I$. 
$I$ is a model of $\Pi$ if it satisfies all rules in $\Pi$. 

For an interpretation
$I$ and a program $\Pi$, the \emph{reduct}
of $\Pi$ w.r.t. $I$ (denoted by $\Pi^I$) is the program
obtained from $\Pi$ by deleting
{\em (i)} each rule $r$ such that $r^- \cap I \neq \emptyset$, and
{\em (ii)} all atoms of the form $\naf a$ in the bodies of the remaining rules.  
Given an interpretation $I$,
observe that the program $\Pi^I$ is a program with no occurrence of $\naf a$.
An interpretation $I$ is an \emph{answer set} \cite{GelfondL90}
of $\Pi$ if $I$ is the least model (wrt. $\subseteq$) of $P^I$.
%


We rely on the notion of an off-line justification, introduced in \cite{pontelli2009justifications}, as an explanation of an atom $a$ given an answer set $A$ of a program $P$. $NANT(P) = \{ a | a \in  r^- \land  r \in P\}$ is the set of all negation atoms in $P$. Let $C(P)$ denote the set of cautious consequences of $P$, i.e., $C(P) = C^+ \cup C^-$  where  $C^+$ is  the set of atoms belonging to all answer sets of $P$ and $C^-$ is the set of atoms which do not belong to any answer set of $P$. A set of atoms $U$ such that $U \subseteq NANT(P) \setminus (A \cup C(P))$ is called a set of assumptions with respect to $A$ if $A = C(P \setminus \{r \in P \mid head(r) \in U\})$. 
\color{black}  
Intuitively, given an answer set $A$ of a program $P$, an atom $a \in A$ ($a \not\in A$) is considered to be true (false) given $A$. An off-line justification for an atom $a$ presents a possible reason for the truth value of $a$, i.e., it answers the question ``\emph{why $a \in A$  (or $\not\in A$)?}''. If $a$ is true in $A$, an off-line justification of $a$ represents a derivation of $a$ from the set of assumptions $U$ and the set of facts in $P$. If $a$ is false in $A$, an off-line justification encodes the reason why it is not supported by $A$, which can be that it is assumed to be false (being an assumption in $U$) or there exists no possible derivation of it given $A$. An off-line justification of $a$ is analogous to the well-known SLDNF tree in that it represents the derivation for $a$. The key difference between these two notions is that an off-line justification might contain a cycle consisting of negative atoms, i.e., atoms not belonging to $A$. For this reason, a justification is represented as an explanation graph, defined as follows.  
%
\color{black} 
Given a program $P$, an answer set $A$, and a set of assumptions $U$ with respect to $A$. 
Let $N = \{x \mid  x \in A\} \cup \{ \sim{x} \mid x \not\in A\} \cup \{\top , \bot, \mathtt{assume}\}$ where $\top$ and $\bot$ represent true and false, respectively. 
An \emph{explanation graph} of an atom $a$ occurring in $P$ is a finite  labeled and directed 
graph $DG_a=(N_a,E_a)$ with $N_a \subseteq  N$ and $E_a \subseteq N_a \times N_a \times  \{+,-,\circ\}$, where $(x,y,z) \in E_a$ 
represents a link from $x$ to $y$ with the label $z$,  and satisfies the following conditions:  
\beitemize 
\item if $a \in A$ then $a \in N_a$ and every node in $N_a$ must be reachable from $a$;  
\item if $a \not\in A$ then $\sim a \in N_a$ and every node in $N_a$ must be reachable from $\sim a$; 
\item if $\{(x, \top,+)\} \in E_a$ then $x$ is a fact in $P$;  
\item if $\{(\sim x, \mathtt{assume}, \circ) \in E_a\}$ then $x \in U$; 
\item if $\{(\sim x, \bot, +) \in E_a\}$ then there exists no rule in $P$ whose head is $x$;
\item there exists no $x,y$ such that $(\top, x, y) \in E_a$, $(\bot, x, y) \in E_a$, or $(\mathtt{assume}, x, y) \in E_a$;
\item for every $x \in N_a \cap A$ and $x$ is not a fact in $P$, 
	\beitemize 
	\item there exists no $y \in N_a \cap A$ such that  $(x,y,-)$ or $(x,y,\circ)$ belongs to $E_a$; 
	\item there exists no $\sim y \in N_a \cap \{\sim u \mid u \not\in A\}$ such that  $(x,\sim y,+)$ or $(x,\sim y,\circ)$ belongs to $E_a$; 
	\item if $X^+ = \{a' \mid (x, a',+) \in E_a\}$ and $X^- = \{a' \mid (x, \sim a',-) \in E_a\}$ then 
	$X^+ \subseteq A$ and $X^- \cap A = \emptyset$ and there is a rule $r$ whose head is $x$ in $P$ such that  
	$r^+ = X^+$ and $r^- = X^-$; and  
	\item $DG_a$ contains no cycle containing $x$.  
	\enitemize 
\item for every $\sim x \in N_a \cap \{\sim u \mid u \not\in A\}$ and $x \not\in U$, 
	\beitemize 
	\item there exists no $y \in N_a \cap A$ such that  $(\sim x,y,+)$ or $(\sim x,y,\circ)$ belongs to $E_a$; 
	\item there exists no $\sim y \in N_a \cap \{\sim u \mid u \not\in A\}$ such that  $(\sim x, \sim y,-)$ or $(\sim x, \sim y,\circ)$ belongs to $E_a$; 
	\item if $X^+ = \{a' \mid (\sim x,   a',-) \in E_a\}$ and $X^- = \{a' \mid (\sim x,  \sim a',+) \in E_a\}$ then 
	$X^+ \subseteq A$ and $X^- \cap A = \emptyset$ and for every rule $r$ whose head is $x$ in $P$ 
	such that $r^+ \cap X^- \ne \emptyset$ or 	$r^- \cap X^+ \ne \emptyset$; and  
	\item any cycle containing $\sim x$ in $DG_a$ contains only node in $N_a \cap \{\sim u \mid u \not\in A\}$.  
	\enitemize 
\enitemize 
\color{black}  
Given an explanation graph $E$ and a node $x$ in $E$, if $x$ is an atom $a$ then the nodes directly connected to $a$---the nodes $y$ such that $(x,y,\_)$ is an edge in $E$---represent a rule whose head is $a$ and whose body is satisfied by $A$; if $x$ is $\sim a$ for some atom $a$,
then the set of nodes directly connected to $a$ represents a set of atoms who truth values in $A$ renders that any rule, whose head is $a$, is unsatisfied by $A$. In other words, the direct connections with a node represents the \emph{support} for the node being in (or not in) the answer set.  
We refer the readers to \cite{pontelli2009justifications} for an in-depth discussion of properties of off-line justifications and the proof of existence of such justifications for every atom in the program.
\color{black}    
Given a program $P$ , an answer set $A$ of $P$, and an atom $a$ occurring in $P$, explanation graphs for $a$ can be generated by (\emph{i}) determining the set of assumptions $U$ with respect to $A$; (\emph{ii}) generating explanation graphs for $a$ using $P$, $A$, and $U$ following its definition.  

\begin{figure}
    \centering
    \includegraphics[width=1\textwidth]{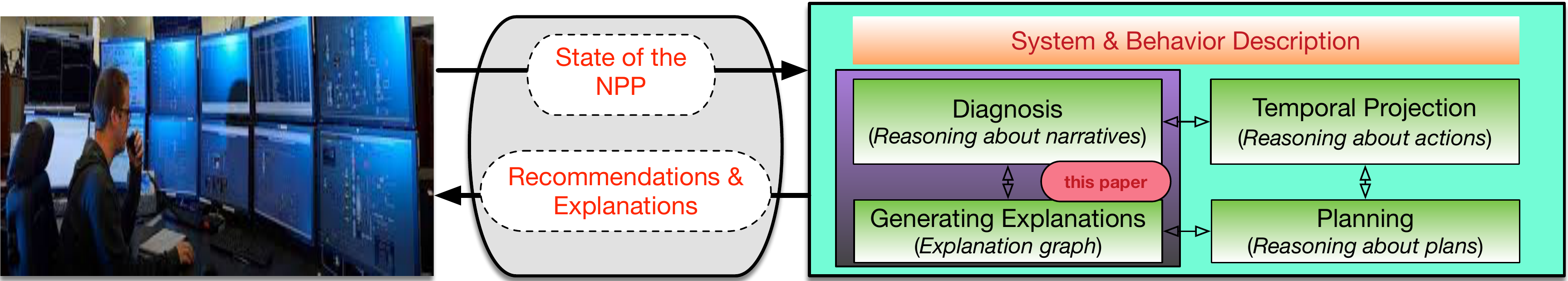}
    \caption{ASP-based operator support system}
  \label{fig:reasoning}
\end{figure}

\section{An ASP-Based Operator Support System for the NPP}
\label{sec:asp-npp}

\begin{figure}
    \centering
    \includegraphics[width=1\textwidth]{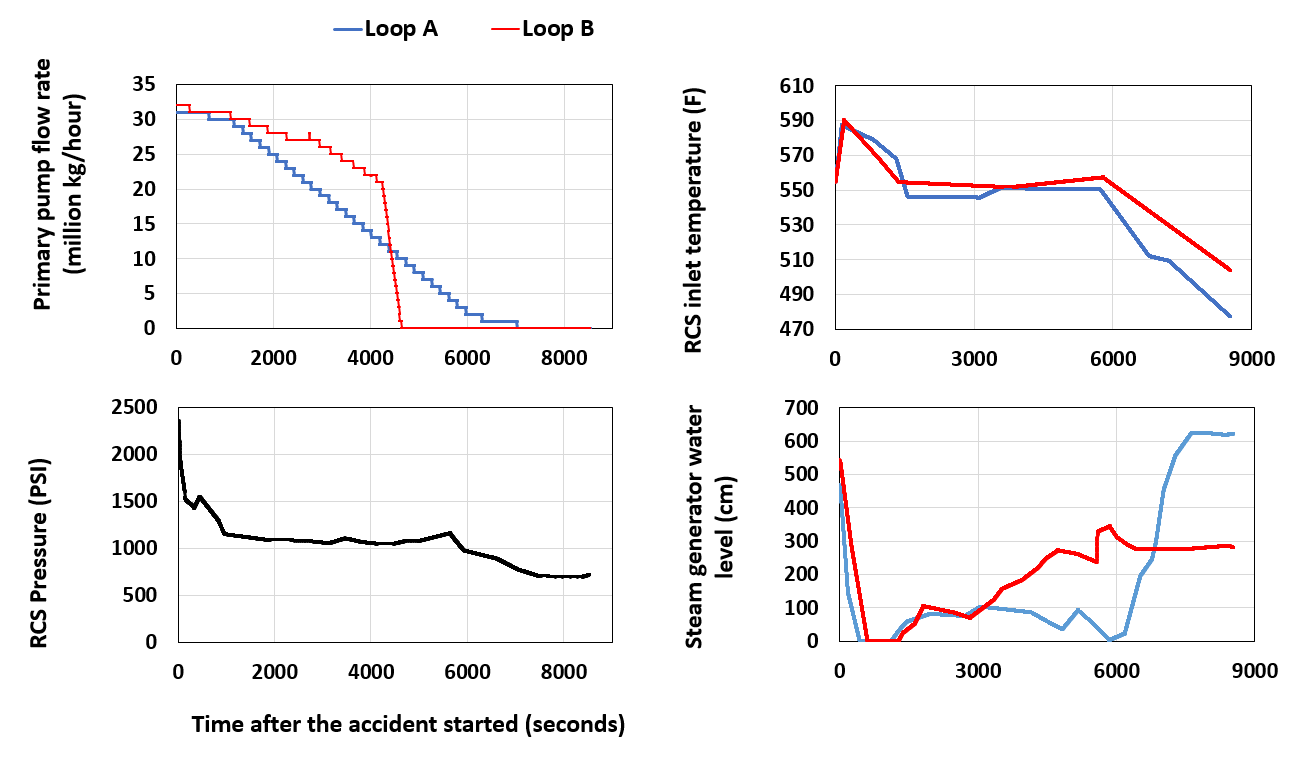}
    \caption{TMI-2 time-dependent variables~\protect\cite{rempe2014instrumentation}.}
  \label{fig:data}
\end{figure}

The overall architecture of the proposed ASP-based operator support system (\asposs) is depicted in Figure \ref{fig:reasoning}. 
It consists of four modules, the temporal projection, planning, diagnosis, and explanation module. The first two modules have been described in detail in \cite{hanna2019artificial}. In this paper, we focus on the diagnosis and explanation modules. 
The diagnosis module is built on the work in using ASP for diagnosis (e.g.,  \cite{balduccini2003diagnostic}) with considerations related to the NPP.  
We use data from the NPP TMI-2 and create inputs (signals) to \asposs{}, simulating the inputs from the NPP to the system.  
These signals carry information about the NPP time-dependent variables and the actions that the operator executed or attempted at any time. For the TMI-2 scenario, the list of relevant time-dependent variables is presented in Table \ref{list}. The first seven variables in Table \ref{list} are plotted in Figure \ref{fig:data}. Due to the lack of data, the remaining nine variables are assigned binary values (for instance, a turbine is on or off ). These 16 variables, were selected, among many measured variables \cite{rempe2014instrumentation} in the NPP, because of their relevance to the TMI-2 scenario and because of the lack of data about other variables.

%

%
%
%
%
\color{black} 
In this work, the NPP system formal description and the rules are inferred from \cite{nuclear1979investigation,nuclear1980analysis}.\color{black} The program encoding the NPP is organized in different sets of rules, which are listed below.  
\beenumerate
\item \underline{Facts}: this set of rules consists of facts declaring the components, variables, the component classification, the ranges of variables, and a steam table to compute the saturation pressure. 
\beenumerate
\item \emph{Components}: 22 components are listed such as\footnote{
   We use {\small\tt clingo}'s syntax. See, e.g., \url{https://potassco.org/clingo/}. 
}:

\begin{table}
\caption{The TMI-2 scenario relevant variables}\label{list}
\begin{tabular}{llll}
\hline\hline
\#     & Variable     & \# & Variable\\
\hline
1   &   Reactor coolant system pressure          &9  &   Condensate pump flow rate (loop B)\\
2   &   Steam generator water level (loop A)        &10  &   Feedwater pump flow rate (loop A)\\
3   &   Steam generator water level (loop B)         &11  &   Feedwater pump flow rate (loop B)\\
4   &    Primary pumps' flow rates (loop A)         &12  &   Emergency feedwater pump flow rate (loop A)\\
5   &   Primary pumps' flow rates (loop B)           &13  &   Emergency feedwater pump flow rate (loop B)\\
6   &   RCS inlet temperature (loop A)              &14 & HPIS pump flow rate\\
7   &   RCS inlet temperature (loop B)              & 15 & Reactor power\\
8   &   Condensate pump flow rate (loop A)          & 16 & Turbine power\\
\hline\hline
\end{tabular}
\end{table}
\begin{verbatim}
component(reactor1; primary_pump_a).
\end{verbatim}
This declares the two components, {\small \tt reactor1} and {\small \tt primary\_pump\_a}. 
\item \emph{Component classification}: Components are divided into groups. Each group have common characteristics, e.g., all valves can be opened or closed and all pumps are characterized by the flow rate. Also, the components that belong to the same loop have the same coolant. For example, the next two rules specify two {\small \tt valve}s and components belonging to the group {\small \tt  primary\_loop\_A\_component}: 
\begin{verbatim}
valve(pressurizer_power_operated_relief_valve;
 	pressurizer_backup_block_valve).
primary_loop_A_component(reactor1; primary_pump_a;
 	steam_generator_primary_a; pressurizer;
 	pressurizer_power_operated_relief_valve;
 	pressurizer_backup_block_valve).	
\end{verbatim}
\item \emph{Ranges of the variables}: e.g., the percentage can have value 0, 1, $\ldots$, 100 and the upper bound of pressure in the primary loop is set to 2255 PSI:  
\begin{verbatim}
percentage(0..100).
upper_pressure_boundary_primary_loop(2255). % PSI
\end{verbatim}
\item \emph{The steam table}: The saturation pressure corresponding to each temperature is encoded by atoms of the form $saturation/2$. If the actual coolant pressure is lower than the coolant saturation pressure, the coolant converts to steam and the reactor coolability is impacted. For instance, if the water temperature is 521F, the water saturation pressure is 820 PSI, as follows:
\begin{verbatim}
saturation(521,820).
\end{verbatim}
\enenumerate
\item \underline{Time-dependent rules}: rules in this group account for variables and actions at previous time steps, include

\beenumerate
\item  \emph{Reasoning about possible actions}: The operator can intervene by executing actions such as increase/decrease the reactor power and increase/decrease the flow rate of any pump.
For each action, an executability condition is defined. An action cannot be recommended unless its execution criteria are satisfied. 
\color{black}  
For instance, the action {\small\tt increase\_control\_rod\_power\_to(Y)} will set the rod power to $Y$ where $Y$ is an integer between 0 and 100,  i.e., {\small\tt rods(C,Y)} will be true. 
This action can be executed at anytime. 
\color{black}  
Seven rules\footnote{To be precise, each rule is about a class of actions. We also omit several ``scenario unrelated actions'' for brevitty.} for actions' execution criteria are implemented. For example, the action of opening an auxiliary feedwater block valve, $aux\_fw\_valve$, is recommended if this valve is closed (as this valve should be normally opened) and the action of opening that block valve is not already attempted. This rule is implemented as:
\begin{verbatim}
recommendation(open,V,T):- aux_fw_valve(V), anytime(T),
   closed(V,T), not attempted(open,V,T1), anytime(T1),
   T<=(T1+M), action_execution_time_range(M).
\end{verbatim}

The status of the valve (open/closed) can be found on the control room panels, and it also can be inferred from other observed variables.

\item \emph{Reasoning about non-observed variables}: Non-observed variables cannot be measured but can be inferred based on the observed variables. Nine rules for inferring non-observed variables are implemented. An example of a non-observed variable is the formation of steam in any loop if the loop pressure is lower than the saturation pressure corresponding to the loop temperature as follows:
\begin{verbatim}
steam(primary_loop_A,T):-
    anytime(T),inlet_temperature_a(X,T),
    primary_loop_pressure(P,T), 
    saturation(X,P1), P<P1.
\end{verbatim}


\item  \emph{Reasoning about triggered actions}: In the NPP, some actions are triggered automatically if certain conditions are met, either by design or by failure of components, and thus might not be executed by the operator. The occurrences of these actions can be inferred by their effects, which can be detected by evaluating the observed variables. Four rules for inferring actions are implemented. For example, the system can infer that a pump is tripped if its flow rate changes from any value above zero to zero. 
\begin{verbatim}
it_happened(trip,P,T):- pump(P), 
    time(T), pump_flow(P,F,T-1), pump_flow(P,0,T), F>0.
\end{verbatim}	

 
\enenumerate
\enenumerate

%


Surprisingly, the program is not very complex. It contains only a few rules with negation as failure atoms (e.g., $\naf \mathtt{attempted}$/3). In total, the module consists of almost 146000 facts and 20 non-ground rules.  The main reason lies in that the history is complete, and our focus is only on analyzing the accident. 

To fully realize the potential of the system in live operation, rules developed for temporal projection and planning modules will be necessary to answer the question of ``\emph{what-if procedure X is executed at time step S?}'' or ``\emph{how can problem Y be addressed?}.'' \color{black} They have been developed in  \cite{hanna2019artificial} based on the action language and reasoning about actions and their effects. The present work shows that the modeling of other aspects in the NPP domain (e.g., components, actions, actions' executability conditions, observed and non-observed variables, static causal laws) could also be benefited from the use of ASP. We note that we could have also added other elements of the action domain described in  \cite{hanna2019artificial}, but they are unnecessary in this setting. This is because values of time-dependent variables (fluents) are already available as real data, and thus rules for action effects on variables would be obsolete for this situation. In addition, because of most fluents in this domain are multi-valued fluents (e.g., a value in the range from 0 to 100), direct encoding of the information related to the NPP in ASP rules is more advantageous than a specification in an action language and then translated into ASP. 
\color{black}

\section{Explaining the TMI-2 Accident} 
\label{sec:asp-exp}
 
\subsection{Diagnosis Module} \label{diagnos_module}
We collect data from \cite{rempe2014instrumentation,nuclear1979investigation} which records the sensor values and the actions that have been executed by the operators during the TMI-2 accident. This data is processed by the \emph{data processing} module and encoded as facts of the form ({\em i}) \emph{variable(value, time)} where \emph{variable} is one of the 16 variables listed in  Table \ref{list} and \emph{time} is in the range [0,8521] (corresponding to 8521 seconds); and (\emph{ii}) \emph{attempted(procedure, component, time)}, which states that the \emph{procedure} is applied on \emph{component} at the \emph{time} (by the operators), and is referred as an attempted action. The list of attempted actions is:
\begin{verbatim}
attempted(open,pressurizer_pilot_operated_relief_valve,7).
attempted(close,pressurizer_pilot_operated_relief_valve,11).
attempted(turn_off,high_pressure_injection_pump,279).
attempted(open,auxiliary_feedwater_a_block_valve,499).
attempted(open,auxiliary_feedwater_b_block_valve,500).
attempted(turn_off,primary_pump_a,4402). 
attempted(turn_off,primary_pump_b,6036).
attempted(close,pressurizer_block_valve,8521).
\end{verbatim}

The~\emph{raw data} are all the variables and actions recorded at all times, from $t_0 = 0$, to $t_{end} = 8521$. To mimic the data streaming in the NPP system, variables, and actions at each timestep, \emph{t}, are extracted, using Python, from the \emph{raw data} file and passed to the \asposs{} system via files. 
\asposs{} processes data within a window of 1 minute, i.e., at the time $t \ge 60$, \asposs{} considers the data in the range $[t-59, t]$. 

For each minute, \asposs{} outputs the computed answer sets, which contain recommendations (e.g., turn on an auxiliary pump), values of inferred variables (e.g., leakage), and the inferred actions (e.g., condensate pump A is tripped). Figure \ref{fig:recommend} illustrates some of the recommendations that the ASP system suggests at specific timesteps. For instance, at the beginning  of the scenario, the auxiliary feedwater pumps were started. Therefore, a recommendation to open the closed auxiliary feedwater line block valve, at $t = 2$, is suggested. Otherwise, the feedwater will be blocked. Because the emergency feedwater line block valve was opened at $t=499$, this recommendation is no longer proposed in the next timesteps. 

Some of the inferred non-observed variables are listed in Figure \ref{fig:nonobs} at specific timesteps. For instance, the lack of water supply (leakage) was inferred at $t = 2$ because of the low steam generator water level. The inferred actions may be similar or different from the attempted actions. The occurrences of these actions are verified by detecting the action effect and watching the observed variables. Figure \ref{fig:infer} lists all the inferred actions at each time step. 


%

\begin{figure}
    \begin{tabular}{ llll }
    \hline 
   Recommended Action &    Recommended Action\\
    \hline\hline 
 \textbf{rec}(open,auxiliary\_feedwater\_a\_block\_valve,\underline{\textbf{2}}).  
& \textbf{rec}(open,auxiliary\_feedwater\_b\_block\_valve,\underline{\textbf{2}}).\\
 ........................................&  ........................................ \\
 \textbf{rec}(close,pressurizer\_backup\_block\_valve,\underline{\textbf{16}}). 
& \textbf{rec}(close,pressurizer\_power\_operated\_relief\_valve,\underline{\textbf{16}}).\\
\textbf{rec}(open,auxiliary\_feedwater\_a\_block\_valve,\underline{\textbf{16}}).
& \textbf{rec}(open,auxiliary\_feedwater\_b\_block\_valve,\underline{\textbf{16}}).\\
 ........................................ & ........................................ \\
 \textbf{rec}(close,pressurizer\_backup\_block\_valve,\underline{\textbf{118}}). 
& \textbf{rec}(close,pressurizer\_power\_operated\_relief\_valve,\underline{\textbf{118}}).\\
 \textbf{rec}(open,auxiliary\_feedwater\_a\_block\_valve,\underline{\textbf{118}}).  
& \textbf{rec}(open,auxiliary\_feedwater\_b\_block\_valve,\underline{\textbf{118}}).\\
 \textbf{rec}(turn\_on,high\_pressure\_injection\_pump,\underline{\textbf{118}}).  
& ........................................\\
 \textbf{rec}(close,pressurizer\_backup\_block\_valve,\underline{\textbf{8521}}). 
& \textbf{rec}(close,pressurizer\_power\_operated\_relief\_valve,\underline{\textbf{8521}}).\\
\textbf{rec}(turn\_on,high\_pressure\_injection\_pump,\underline{\textbf{8521}}). &  \\
    \hline
    \end{tabular}
\caption{Output: Recommendations (\textbf{rec} stands for recommendation, \underline{\textbf{x}} stands for the time step $x$ )}
\label{fig:recommend}
\end{figure}

\begin{figure}
    \begin{tabular}{ l l}
    \hline
Inferred non-observed variables  & Inferred non-observed variables \\ 
 \hline \hline    
closed(auxiliary\_feedwater\_a\_block\_valve,\underline{\textbf{2}}).
& closed(auxiliary\_feedwater\_b\_block\_valve,\underline{\textbf{2}}). \\    
lack\_of\_water\_supply(secondary\_loop\_A,\underline{\textbf{2}}).
    & lack\_of\_water\_supply(secondary\_loop\_B,\underline{\textbf{2}}).\\
closed(auxiliary\_feedwater\_a\_block\_valve,\underline{\textbf{18}}).& 
closed(auxiliary\_feedwater\_b\_block\_valve,\underline{\textbf{18}}).   \\ 
 lack\_of\_water\_supply(secondary\_loop\_A,\underline{\textbf{18}}). & 
lack\_of\_water\_supply(secondary\_loop\_B,\underline{\textbf{18}}).\\
 stuck\_open(\textbf{pporv},18). 
& ........................................\\
closed(auxiliary\_feedwater\_a\_block\_valve,\underline{\textbf{847}}).
& closed(auxiliary\_feedwater\_b\_block\_valve,\underline{\textbf{847}}).\\
 lack\_of\_water\_supply(secondary\_loop\_A,\underline{\textbf{847}}). 
 & lack\_of\_water\_supply(secondary\_loop\_B,\underline{\textbf{847}}).\\
 steam(primary\_loop\_A,\underline{\textbf{847}}).
& stuck\_open(\textbf{pporv},\underline{\textbf{847}}).\\
 ........................................
& stuck\_open(\textbf{pporv},\underline{\textbf{8521}}).\\
    \hline
    \end{tabular}
\caption{Output: non-observed variables (\textbf{pporv}: pressurizer\_power\_operated\_relief\_valve)}
\label{fig:nonobs}
\end{figure}

We observe that this module of \asposs{} would have provided useful assistance to the operator in the decision making process for the following reasons:
\beenumerate
\item The operators \emph{did not notice} that the primary loop water reached saturation pressure (water converted to steam causing inadequate reactor cooling). The existence of steam is inferred by \asposs{} (see Figure \ref{fig:nonobs}, at $t=847$).
\item Because of the reasoning system capacity to monitor more variables simultaneously, the reasoning system gives \emph{early} recommendations for closing the pressurizer block valve (see Figure \ref{fig:recommend} at $t=16$) and opening the emergency feedwater line block valves (see Figure \ref{fig:recommend} at $t=2$). In reality, these decisions have been \emph{delayed} because the pressurizer block valve was closed at $t= 8521$, and the emergency feedwater line block valves were opened at $t =499$ (see Table \ref{events}).

%

\item \asposs{} could distinguish between the executed action, \emph{$it\_happened/3$} in Figure \ref{fig:infer}, and the attempted actions, \emph{$attempted/3$}, by watching the action effect. This feature is significant for the TMI-2 scenario because of the confusion about the PORV. An action to close the PORV was attempted at $t=11$ (see the attempted actions in Section \ref{diagnos_module}), but the valve was not closed (see the executed actions in Figure \ref{fig:infer}), and the primary loop pressure kept decreasing.
%
%
%
\enenumerate

\begin{figure}
    \begin{tabular}{ l l l l }
    \hline
 Time & Inferred Actions & & \\ 
 \hline \hline    
%
1 & \textbf{ith}(trip,condensate\_pump\_a,1). & 1 
    & \textbf{ith}(trip,condensate\_pump\_b,1). \\
2  & \textbf{ith}(trip,feedwater\_pump\_a,2).  & 2  
    & \textbf{ith}(trip,feedwater\_pump\_b,2). \\
2    & \textbf{ith}(trip,turbine1,2).  & 2 
    & \textbf{ith}(start,auxiliary\_feedwater\_pump\_a,2). \\
2    & \textbf{ith}(start,auxiliary\_feedwater\_pump\_b,2). &  
11  & \textbf{ith}(trip,reactor1,11). \\
122 & \textbf{ith}(start,high\_pressure\_injection\_pump,122). & 
278 & \textbf{ith}(trip,high\_pressure\_injection\_pump,278). \\
4403 & \textbf{ith}(trip,primary\_pump\_a,4403). &
6037 & \textbf{ith}(trip,primary\_pump\_b,6037). \\
    \hline
    \end{tabular}
\caption{Output: inferred actions ($\mathbf{ith}$ stands for \emph{it\_happened})}
\label{fig:infer}
\end{figure}

%
%

\subsection{Generating Explanations} 
 
\begin{figure}
  \centering
  \includegraphics[width=0.7\textwidth]{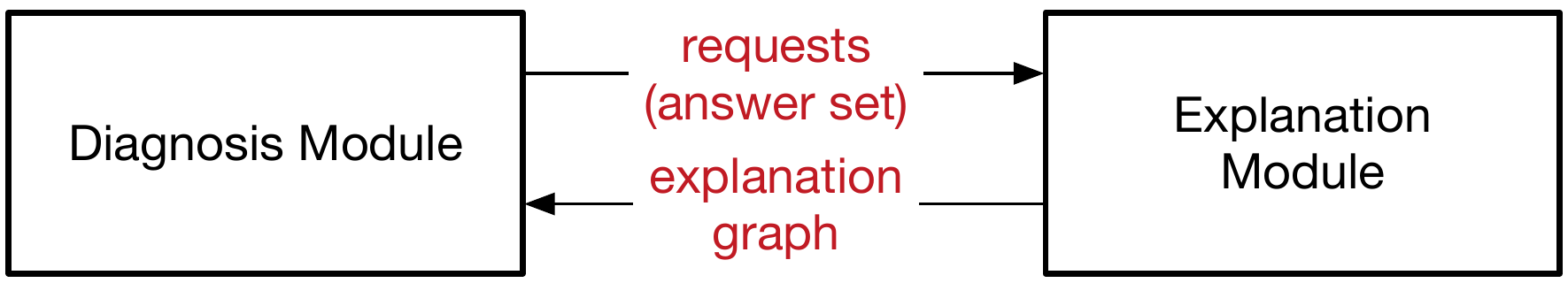}
\caption{Communication Between Diagnosis and Explanation Modules \label{fig:explanation_module}}  
\end{figure}

We implemented an explanation module for generating explanations for the answers computed by the diagnosis module.
In this experiment, these two modules run in parallel as two cooperating processes (written in Python) and interact with each other via messages (Figure~\ref{fig:explanation_module}). The explanation module has two threads, one for receiving requests from the diagnosis module and one for returning the explanation graphs. Whenever a request from the diagnosis module arrives, the receiving thread puts it into a queue. Once the explanation module recognizes that a request arrived (the queue is not empty), it computes the explanation graphs (for atoms indicating the recommended actions and the inferred variables/actions) and displays them. We note that the focus is on these atoms because they are the most significant atoms in the answer sets, reflecting the activities within the nuclear reactor. This can be relaxed by allowing the diagnosis module to indicate the atoms of interest in the request.  

\begin{figure}
  \centering
  \includegraphics[width=1\textwidth]{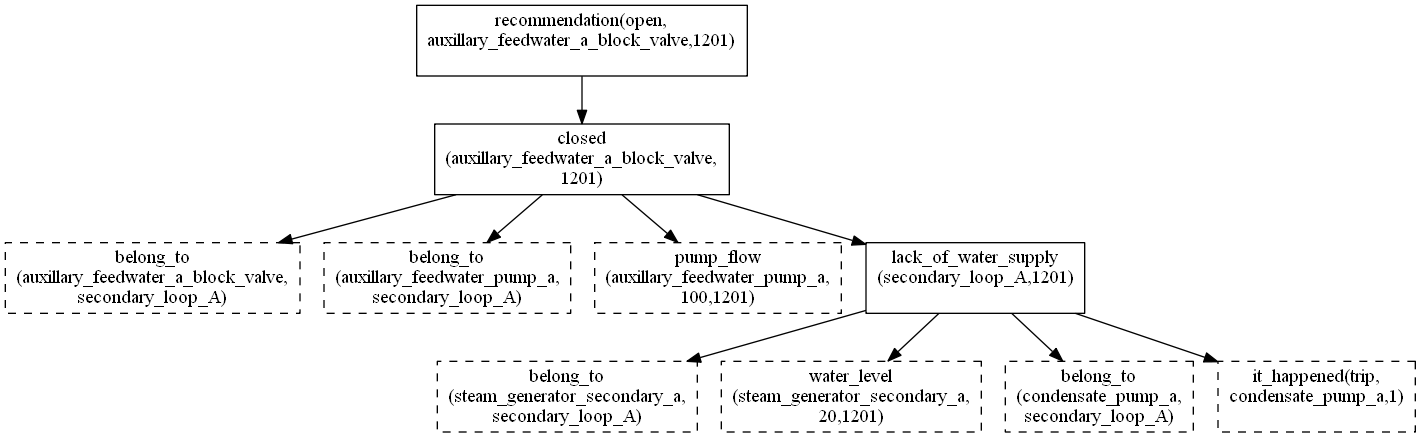}
  \caption{The first explanation of $recommendation(open,auxiliary\_feedwater\_a\_block\_valve,1201)$}
  \label{fig:recommendation1}
\end{figure}

\begin{figure}
  \centering
  \includegraphics[width=1\textwidth]{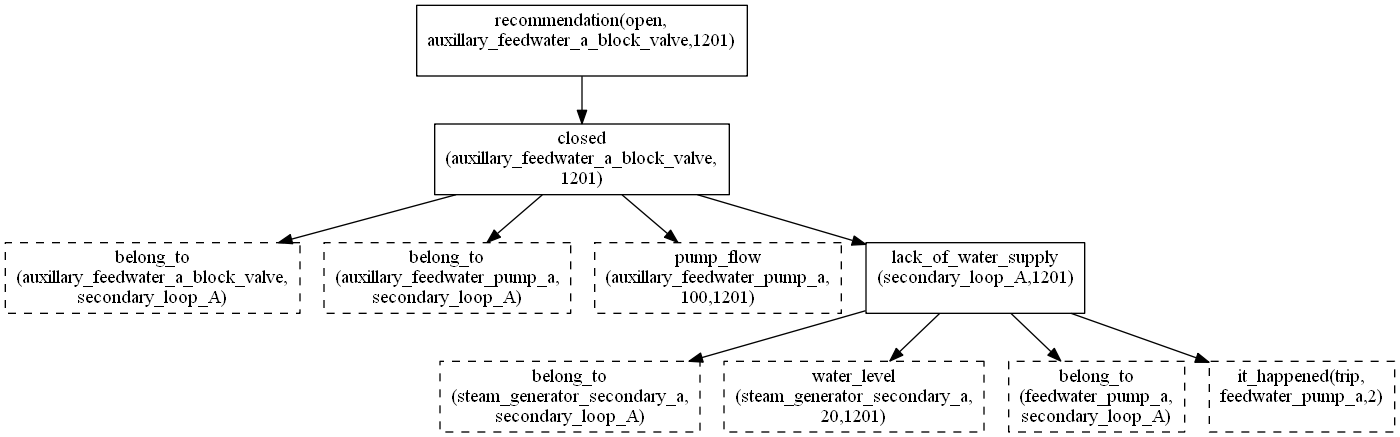}
  \caption{The second explanation of $recommendation(open,auxiliary\_feedwater\_a\_block\_valve,1201)$}
  \label{fig:recommendation2}
\end{figure}

The explanation module computes and draws the explanation graph as defined in Subsection~\ref{subsection:asp} for atoms occurring in \asposs. \color{black} The explanation module provides all possible explanation graphs for the chosen atoms (e.g. $recommendation/3$ in Figure~\ref{fig:recommendation1} and Figure~\ref{fig:recommendation2}) to support the decision making for the operators\footnote{ 
Observes that explanation graphs (or justifications) could be compared with each other (e.g., via a preference relation). As such, we could allow operators to specify their preferences and the system returns the set of \emph{preferred} explanation graphs . Dealing with preferences of this type can be done straightforwardly in ASP. However, anecdotal stories related to TMI-2 suggest that being able to present \emph{all possible explanations} is important. Indeed, no one would ever think of the possibility that the valve, PORV, was stuck open. Preferences are often given to components deemed more ``important''  than a valve. It was said that if someone has noticed this problem on time and closed the block valve, the accident would not have happened. Indeed, this is one of the actions recommended by the system very early on (see the recommended action at $time= 16$, Figure~\ref{fig:recommend}). For this reason, our focus in this paper was on generating all explanations. 
}.


\color{black} 
We use solid, dash and dot lines to represent $+$, $-$ and $\circ$ links, respectively. In all graphs, we also omit all links to $\top$ from nodes with 
dash boxes. Some examples are listed below: 
\beitemize 

\begin{figure} 
  \centering
  \includegraphics[width=.4\textwidth]{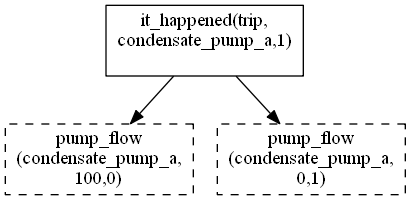}
  \includegraphics[width=.5\textwidth]{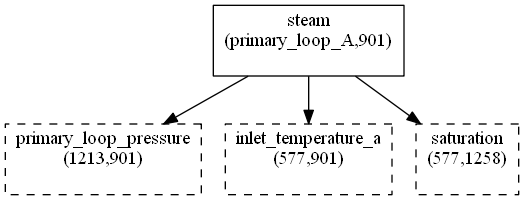}
  \caption{Left: Explanation of $it\_happened(trip,condensate\_pump\_a,1)$. Right: Explanation of $steam(primary\_loop\_A,901)$}
  \label{fig:happened}\label{fig:steam}
\end{figure}

\item Figure~\ref{fig:happened} (left) shows that the condensate pump A is tripped at $t=1$ because its pump flow at $t=0$ is $100$, and it sharply decreases to $0$ at $t=1$.

\item Figure~\ref{fig:steam} (right) indicates the formation of steam in the primary loop $A$, at $t=901$ because the pressure of the primary loop A ($1213~PSI$) is less than the saturation pressure corresponding to the inlet temperature of the primary loop A ($1258~PSI$).

\item \color{black} The auxiliary feedwater line block valve (loop A) opened at $t=1201$  has two possible explanations which are presented in Figure~\ref{fig:recommendation1} and Figure~\ref{fig:recommendation2}. The recommendation is to open the auxiliary feedwater line block valve because the valve is closed, and there are no attempted actions to open that block valve. The valve is closed because a lack of water supply, in the secondary loop, at $t=1201$ is detected. Lack of water supply is inferred, in the secondary loop, because the condensate pump in that secondary loop is tripped at $t = 1$ (Figure~\ref{fig:recommendation1}) or the feedwater pump in that secondary loop is tripped at $t = 2$ (Figure~\ref{fig:recommendation2}), and the corresponding steam generator water level ($20~cm$) is below a minimum value ($30~cm$ by default). \color{black}

%
%
\enitemize
\color{black} 
\subsection{Performance}  
In this work, the computational expense of the overall ASP-based reasoning system, accounting for 16 variables and eight actions within 8521 seconds (142 minutes) of the accident scenario, is 30 minutes (on a four-processor machine). It was assumed that the values of all variables were updated each second, and new outputs are computed each second. Depending on the variables' time scales, answers can be generated at a slower rate to decrease the ASP computational time.
We found that the computation of the explanation module would not be a bottleneck of the overall system because it takes at most 5 seconds to compute the first batch of the explanation graphs. 

We use {\small \tt clingo} as the off-the-shelf tool in this development. Because of {\tt\small clingo} simplified rules and false atoms (e.g., atoms which do not occur in the head of any rules are removed from the program), the explanation graphs are computed using the simplified programs from {\small\tt clingo}. As such, some theoretical graphs/links for certain atoms might not be returned from the explanation module.\color{black} 

\color{black} 
Practically, implementing an AI-guided decision support system still faces some challenges as many NPPs rely on traditional analog technology. Also, relying on AI may exacerbate the concerns about NPP cybersecurity. Nevertheless, the explainability of the ASP reasoning system outputs gives ASP an advantage over other methods (e.g., machine learning methods). 
\color{black} 

%
%
%
%
%

\section{Conclusions}

We discussed an application of ASP in analyzing the most severe nuclear accident in US history, the Three Mile Island Unit 2 (TMI-2) accident scenario,  focusing on the ability of reasoning about an event history to identify diagnoses and generating explanations of ASP. The experiment shows that the ASP-based system can assist NPP operators in several ways: 
\beenumerate
\item \emph{Augmented capabilities}: It provides operators with the capability to reason with several variables and monitor many dynamic events at the same time. 
In the TMI-2 scenario, the operator was overwhelmed by many alarms and monitoring many dynamic events. In our experiment, the system needs only around 30 minutes to process the information of the first 142 minutes of the scenario, identifying actions that should have been executed much earlier than the operators (see Subsection~\ref{diagnos_module}).
\item \emph{Explanation}: It provides explanations for the recommendations or information related to the event history (e.g., inferred values or inferred actions) that the operators would not have been able to observe.  
\enenumerate
This work shows that ASP, and more generally, declarative programming can be employed in supporting the control of complex and critical systems.

In the near future, we would like to integrate the modules proposed in this paper with the reasoning system in \cite{hanna2019artificial} to provide a platform for supporting NPP operators which can take into consideration the uncertainty associated with actions and observations (e.g., sensor values) as well as, potentially, users' preference and knowledge in planning and selecting a smaller set of explanations. 

While we focus on generating all explanations in this paper,  it might be useful to provide the users with choices (e.g., use vs. not use preference ). Last but not least, we would like to identify the opportunity to work with NPP operators, by providing the system as a ``passive supporter,''  which monitors the environment and provides users with explanations whenever it is asked to do so.  
\color{black} 

%
%

\section{Acknowledgments}
This research is supported by the US DoE's  Advanced Research Project Agency-Energy (ARPA-E) MEITNER Program through award DE-AR0000976. 
The authors would like to thank Dr. Robert W. Youngblood III (Idaho National Laboratory)for his constructive comments on earlier versions of this paper and his continuous support throughout the development of an ASP-based system for NPP operators.  
The third author is also partially supported by the NSF-grant \#1812628, \#1914635, and \#1757207.

\bibliographystyle{acmtrans}
\bibliography{bibliography} %

\label{lastpage}
\end{document}